\documentclass[preprint,12pt]{elsarticle}

\usepackage{amsmath}
\usepackage{amssymb}
\usepackage{amsfonts}
\usepackage{graphicx}
\usepackage{textcomp}
\usepackage{xcolor}
\usepackage{bm}
\usepackage{booktabs}
\usepackage{array}
\usepackage{caption}
\usepackage{dblfloatfix}
\usepackage{multirow}
\usepackage{booktabs}
\usepackage{pifont}
\usepackage{bbding}
\usepackage{hyperref}
\usepackage{algorithm}
\usepackage{algpseudocode}

\usepackage[switch]{lineno}
\usepackage{mmstyles}

\begin{document}

\begin{frontmatter}


\title{X-ray Insights Unleashed: Pioneering the Enhancement of Multi-Label Long-Tail Data}


\author[label2]{Xinquan Yang}
\ead{xinquanyang99@gmail.com}
\author[label3]{Jinheng Xie}
\ead{xiejinheng2020@email.szu.edu.cn}
\author[label1]{Yawen Huang\corref{mycorrespondingauthor}}
\ead{yawenhuang@tencent.com}
\author[label1]{Yuexiang Li}
\ead{yuexiang.li@ieee.org}
\author[label1]{Huimin Huang}
\ead{huiminhuang@tencent.com}
\author[label1]{Hao Zheng}
\ead{howzheng@tencent.com}
\author[label1]{Xian Wu}
\ead{kevinxwu@tencent.com}
\author[label1,label4]{Yefeng Zheng}
\ead{zhengyefeng@westlake.edu.cn}
\author[label2]{Linlin~Shen\corref{mycorrespondingauthor}}
\ead{llshen@szu.edu.cn}

\cortext[mycorrespondingauthor]{Corresponding author}

\address[label1]{Tencent Jarvis Lab, Shenzhen, China}
\address[label2]{School of Artificial Intelligence, Shenzhen University, Shenzhen, China}
\address[label3]{National University of Singapore, Singapore}
\address[label4]{Westlake University, Hangzhou, China}

\begin{abstract}

Long-tailed pulmonary anomalies in chest radiography present formidable diagnostic challenges. 
Despite the recent strides in diffusion-based methods for enhancing the representation of lesions of tail classes, the paucity of rare lesion exemplars curtails the generative capabilities of these approaches, thereby leaving the diagnostic precision less than optimal. 
In this paper, we propose a novel data synthesis pipeline designed to augment tail-class lesions utilizing a copious supply of conventional normal X-rays. 
Specifically, a sufficient quantity of normal samples is amassed to train a diffusion model capable of generating normal X-ray images. 
This pre-trained diffusion model is subsequently utilized to inpaint the head-class lesions present in the diseased X-rays, thereby preserving the tail classes as augmented training data.
Additionally, we propose the integration of a Large Language Model Knowledge Guidance (LKG) module alongside a Progressive Incremental Learning (PIL) strategy to stabilize the inpainting fine-tuning process. 
Comprehensive evaluations conducted on the public lung datasets, i.e., MIMIC and CheXpert, demonstrate that the proposed method sets a new benchmark in performance.
\end{abstract}


\end{frontmatter}

\section{Introduction}\label{sec1}
Chest X-ray (CXR), as the most common technique in the realm of medical imaging, plays a pivotal role in the diagnosis of thoracic diseases. 
Despite the rapidity and convenience of CXR imaging, the onus falls upon radiologists to manually scrutinize thousands of radiological samples daily, a task that is both arduous and burdensome. 
With the advent of deep learning paradigms and the burgeoning availability of CXR datasets~\cite{bustos2020padchest,johnson2019mimi,nguyen2022vindr,irvin2019chexpert,wang2017chestx}, significant strides have been made in the realm of CXR diagnostics, particularly in disease detection~\cite{haque2024robust}, automated report generation~\cite{liu2021auto}, and CXR image synthesis~\cite{bluethgen2024vision}.

Typically, a single CXR image may contain multiple thoracic diseases, necessitating the development of multi-label CXR classification, wherein the classification network yields multiple disease categories. 
However, the presence of rare diseases (tail classes) in CXR images often results in suboptimal performance by the classification networks, which predominantly favor more common thoracic diseases (head classes) and tend to neglect the distinctive features of these less prevalent categories. 
This disparity manifests as the long-tail (LT) problem. 
As shown in Figure~\ref{fig_distribute}(a), this long-tail distribution is obvious across multiple datasets, characterized by a preponderance of data in the head classes and a scarcity in the tail classes.

\begin{figure}
\centering
\includegraphics[width=0.8\linewidth]{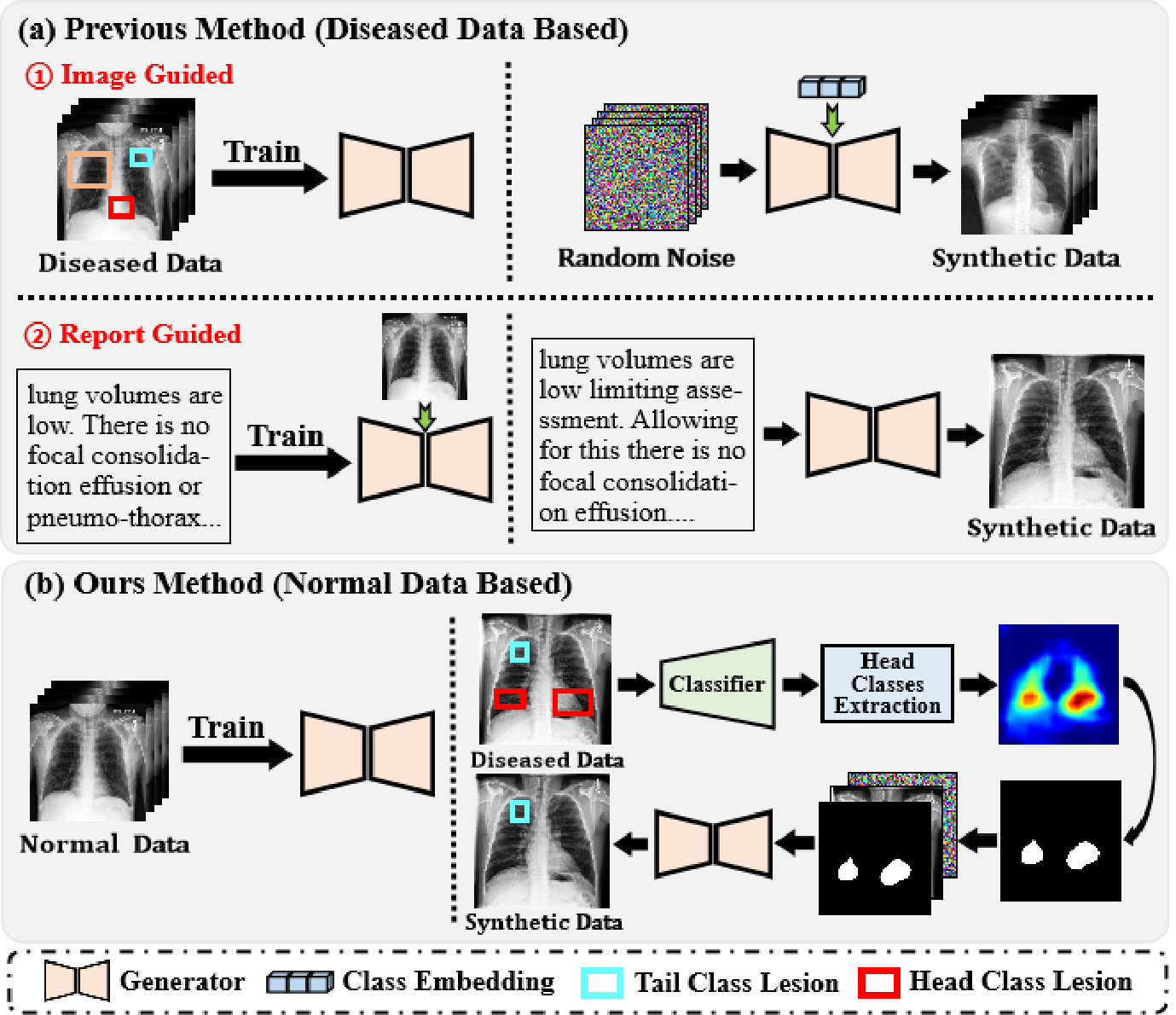}
\caption{Comparison between previous Chest X-ray data augmentation and the proposed method. (a) Previous method based on diseased data for generation.
(b) Our method based on normal data for inpainting.} \label{fig1}
\end{figure}

\begin{figure*}
\centering
\includegraphics[width=1\linewidth]{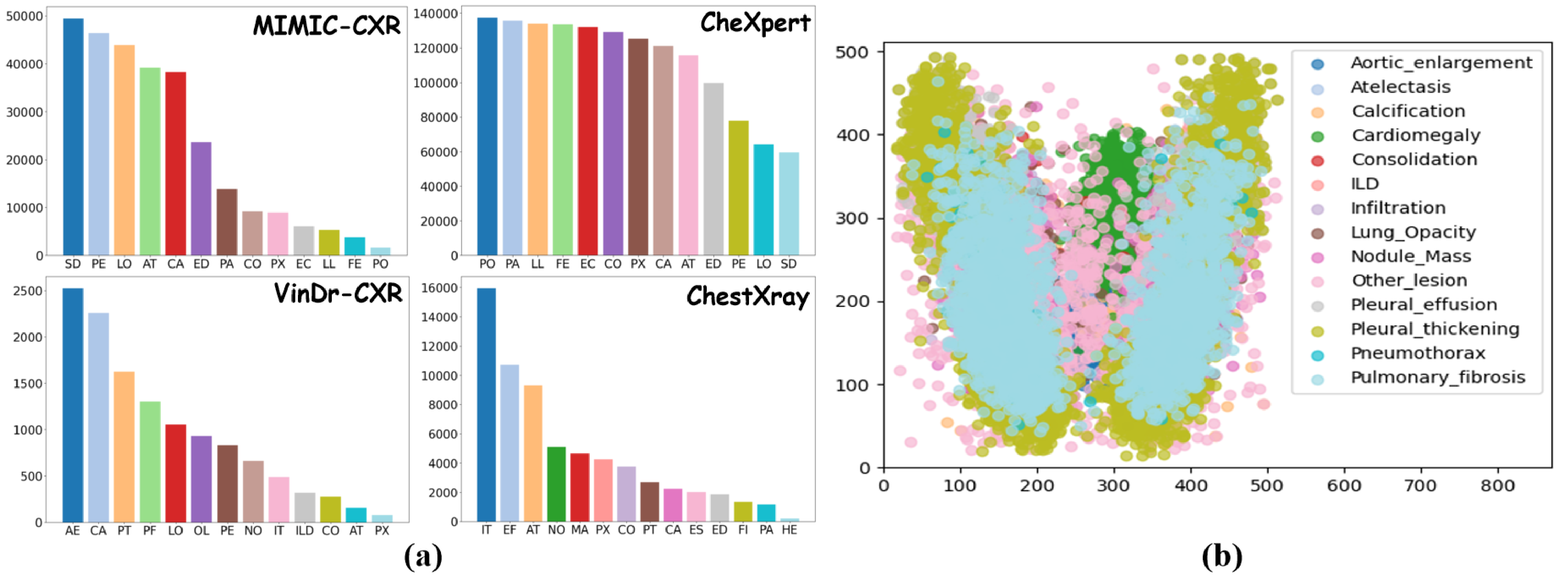}
\caption{(a) The data distribution in each public dataset. (b) Diagram of lung disease entanglement. Each point in the figure is the center of the lesion annotation in the VinDr-CXR~\cite{nguyen2022vindr} dataset.} \label{fig_distribute}
\end{figure*}

Researchers have endeavored to address the long-tail problem in medical imaging through non-generative methods, which require additional statistical support, such as the computation of data distribution statistics, or involve labor-intensive procedures, including fine-tuning and the manual delineation of head-tail class divisions. 
Typical strategies encompass data re-sampling and loss re-weighting. 
Data re-sampling methods~\cite{gupta2019lvis} demand pre-computed data distributions, potentially overfitting in tail classes and underfitting in head classes. 
Similarly, loss re-weighting approaches~\cite{cui2019class, kang2019decoupling, li2020overcoming} depend on these statistics to adjust weights, elevating the importance of tail classes while diminishing that of head classes at the class level. 
Both methods are heavily relied on complex computations of prior data, which may subsequently constrain the enhancement of performance for tail classes.


Recently, researchers have explored the potential of training a CXR image generator to facilitate data augmentation. 
As shown in Figure~\ref{fig1}(a), the CXR generator produces images based on either textual report descriptions~\cite{bluethgen2024vision} or stochastic noise inputs~\cite{shentu2024cxr}. 
However, this approach encounters challenges in addressing tail classes, as the training data for the generative model also exhibits a long-tail distribution. 
It becomes evident that training a generator capable of reliably producing high-quality data for tail classes is an impractical expectation. 
Contrary to tail class data, the quantity of normal samples in clinical practice significantly outnumbers that of abnormal samples.
This observation inspires us to consider:
\begingroup
\addtolength\leftmargini{-0.05in}
\begin{quote}
     \textit{Can \textbf{normal X-rays} be used for data augmentation of tail classes?}
\end{quote}
\endgroup

Intuitively, the proposition is to train a normal CXR generator that inpaints regions of easily detected lesions (e.g., head classes) to restore normal lung texture, while the remaining challenging-to-detect categories (e.g., tail classes) constitute a new training image. 
As depicted in Figure~\ref {fig1}(b), we initially train a normal X-ray diffusion model and a lung lesion classifier. 
The classifier identifies the disease areas of head classes using a class activation map (CAM)~\cite{selvaraju2017grad}. 
The CAM map is then transformed into a mask and inputted into the pre-trained CXR diffusion model to generate a new image. 
Through this process, the head classes in the image are eliminated, while the tail classes persist and become new training data, thereby augmenting the number of tail samples. 
However, this process encounters several challenges: \textbf{Disease Entanglement}: The imaging characteristics of chest X-rays may cause different disease categories to become entangled (as shown in Figure~\ref{fig_distribute}(b)). 
When head and tail classes appear in the same area, inpainting the head class may lead to incorrect data generation. 
\textbf{Domain Gap}: Given that the CXR generator is trained on disparate datasets, the domain of the inpainted image may diverge from that of the original dataset, leading to a decline in model performance.

To address these challenges, we propose a Large Language Model Knowledge Guidance Module (LKG) and a Progressive Incremental Learning (PIL) strategy. 
The LKG leverages the lung-related knowledge embedded within the LLM to determine whether the disease class in the current image is entangled. 
If entanglement is detected, the LKG filters out the entangled classes to ensure the accuracy of the inpainting process. 
Concurrently, the PIL employs a progressive training method to incrementally add inpainted images to the training set based on the number of training epochs. This method circumvents catastrophic forgetting of head-class data when incorporating a large volume of tail-class data for training. 
As a result, it ensures that the performance of the head classes is preserved while simultaneously enhancing the performance of the tail classes.

In summary, our main contributions are as follows: 
\begin{itemize} 
\item To the best of our knowledge, this is the first work to leverage normal X-rays to enhance the performance of tail classes, which is more practical for clinical application. 
\item We release a normal chest X-ray generator trained with a large number of diverse source datasets. It possesses robust generation capabilities and can adapt to the inpainting of different source X-rays. 
\item A large language model knowledge guidance module is designed to leverage the lung-related knowledge embedded within the LLM to resolve the problem of disease entanglement.
\item A progressive incremental learning strategy is proposed to stabilize the training of different domain images. 
\item Extensive experiments on two public lung datasets (MIMIC-CXR and CheXpert) demonstrate that the proposed method can effectively improve the performance of tail classes in real datasets.
\end{itemize} 

\section{Related work}\label{sec2}
\subsection{Lung Disease Recognition}
With the development of deep learning, researchers have proposed many methods for automatic disease recognition in chest X-ray (CXR) images. Cervantes et al.~\cite{cervantes2021lime} proposed a simple and effective approach for COVID-19 screening by analyzing CXR images using CNNs. The study utilized four pre-trained CNN networks from ImageNet: ResNet50, VGG16, DenseNet201 and EfficientNetB3, for finetuning on a dataset containing CXR images of COVID-19 cases, healthy individuals, and patients with viral pneumonia.
Priya et al.~\cite{mohanapriya2019lung} devised a lung diseases recognization system relying on deep convolutional neural networks, emphasizing the potential of CNNs in detecting lung abnormalities such as cancer. This study highlighted the importance of transfer learning, which enables leveraging pretrained CNN architectures for smaller datasets, thus contributing to the success of such studies. 
Abdul et al.~\cite{abdul2020automatic} introduced an automatic lung cancer detection and classification system designed to solve the challenge of early lung cancer detection using computed tomography images. 

Afterward, some works applied Vision Transformer (ViT)~\cite{dosovitskiy2020image} to improve the visual feature extraction in the chest X-rays. Chen et al.~\cite{chen2024vision} analyzed and compared the performance of four diverse architectures: ViT, EfficientNet~\cite{tan2019efficientnet}, MViT~\cite{li2022mvitv2}, and EfficientViT~\cite{liu2023efficientvit}, to explore the influence of varying the number of encoder blocks within the transformer models. Rocha et al.~\cite{rocha2024stern} proposed STERN, composed of an attention-driven spatial transformer module and a classifier, which is able to select the thoracic region of interest, transform the input image accordingly, and then classify it as normal or abnormal. Zhang et al.~\cite{zhang2024three}  proposed a three-stage framework with knowledge transfer from adult chest X-rays to aid the diagnosis and interpretation of pediatric thorax diseases. 

\subsection{Long Tail Data Augmentation}
Researchers have proposed many solutions for long-tail datasets, which can be divided into generative and non-generative. Representatives of non-generative solutions are data augmentation and loss reweighting. 
Galdran et al.~\cite{galdran2021balanced} proposed Balanced-MixUp, an extension of the MixUp~\cite{zhang2017mixup} regularization technique with class-balanced sampling, a general augmentation approach in the long-tail learning. 
Ju et al.~\cite{ju2021relational} classified tail classes into subsets based on prior information (clinical presentation and location) and used knowledge distillation to train a teacher model to enforce the student model to learn these prior knowledge. 
Zhang et al.~\cite{zhang2021mbnm} combined a feature memory module, resampling of tail classes, and a re-weighted loss function to improve the long-tail classification performance. 
Kang et al.~\cite{kang2019decoupling} proposed to decouple the learning procedure into representation learning and classification, and systematically explore how different balancing strategies affect for long-tailed recognition. 
Li et al.~\cite{li2020overcoming} proposed a novel balanced group softmax module for balancing the classifiers within the detection frameworks through group-wise training. 
Kozerawski et al.~\cite{kozerawski2020blt} proposed a data augmentation technique that compensates the imbalance of long-tailed classes by generating hard examples via gradient ascent techniques from existing tail-class training examples.

The main idea of the generative method is to increase the amount of data in the tail classes. Shen et al.~\cite{shen2023image} proposed a novel lung nodule synthesis framework based on image inpainting, which hierarchically decomposes the nodule synthesis procedure into shape generation, size modulation, and texture synthesis. Tang et al.~\cite{tang2021disentangled} proposed a deep disentangled generative model simultaneously generating abnormal disease residue maps and “radiorealistic” normal CXR images from an input abnormal CXR image. Luo et al.~\cite{luo2024measurement} devised a new diffusion synthesis framework, which introduces uncertainty guidance in each sampling step and design an uncertainty-guided diffusion models. Tang et al.~\cite{tang2024text} introduced a novel text-guided tail-class generation network, which first learns the feature representations of from each other via knowledge mutual distillation network, thereby enhancing the feature extraction capability of the tail-class model. 

Different from these methods, we propose to train a normal X-ray diffusion model to inpaint abnormal head-class regions while retaining the tail regions, thereby constructing new tail-class data. Since normal X-rays are readily available and abundant, our method can be applied to arbitrary datasets rather than specific disease categories.
 
\begin{figure*}[t!]
\centering
\includegraphics[width=1\linewidth]{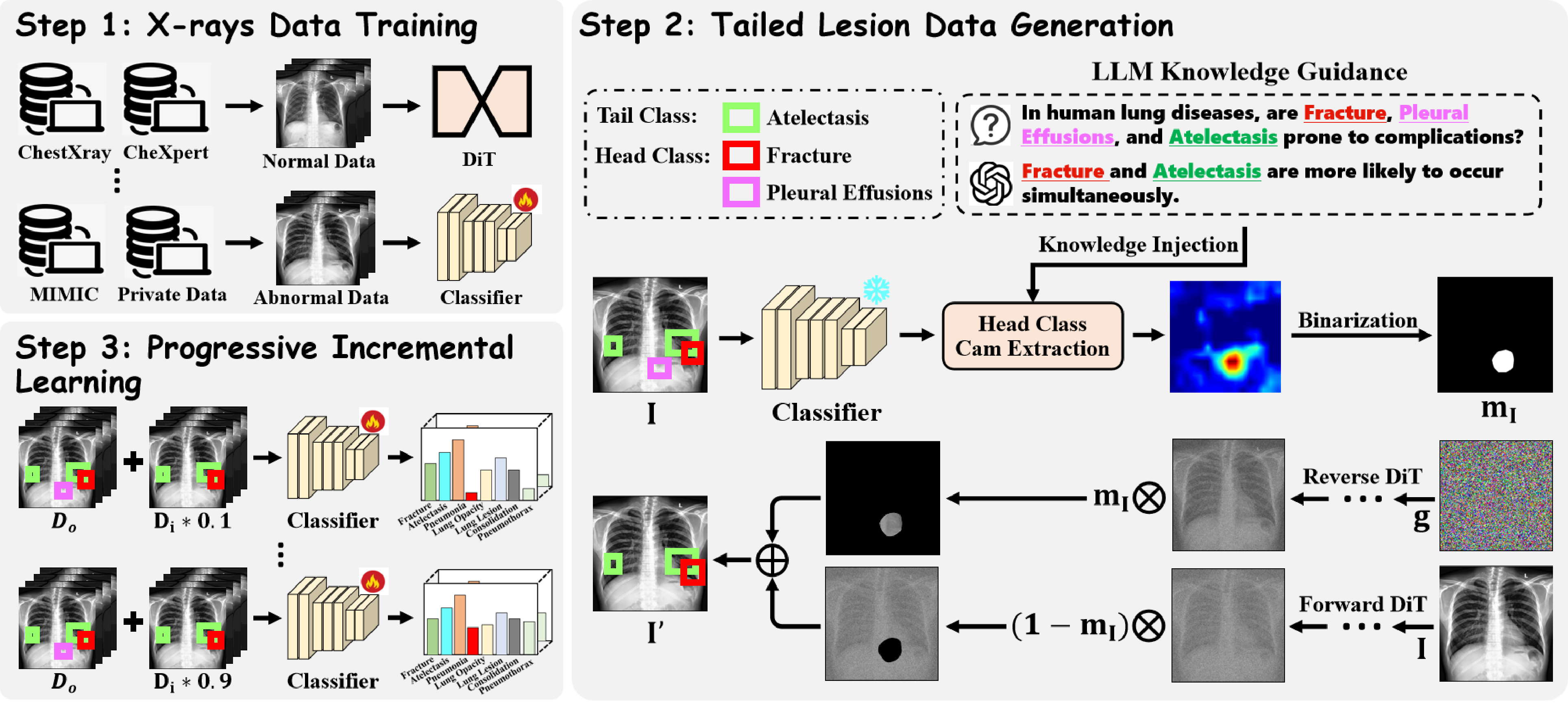}
\caption{Overview of our proposed framework.} \label{fig_network}
\end{figure*}

\section{Method}
Figure~\ref{fig_network} provides an overview of the proposed method, which comprises three main steps: (1) Data collection and model pre-training: Normal and abnormal data are collected from public and private datasets. The normal data is utilized to train a Diffusion Transformer (DiT) model, while the abnormal data is used to train an initial classifier. (2) Tail-class lesion data generation: Class activation mapping (CAM) is applied to the classifier to identify head-class lesion regions. The pre-trained DiT model is then employed to restore these lesion areas in the original image to normal lung texture, thereby generating new tail-class lesions data. (3) Progressive model fine-tuning: The classifier is fine-tuned using a progressive incremental learning (PIL) strategy, which involves gradually incorporating the newly generated tail-class lesion data into the original training set for mixed training. Each of these steps is elaborated in detail in the following sections. 

\subsection{Data Collection and Model Pre-training}
\subsubsection{DiT Training on Normal X-rays}
Our objective is to inpaint the lesion regions belonging to head classes in chest X-rays with normal lung textures, while preserving the lesions of tail classes. 
This requires a robust generator for normal X-ray images. 
To this end, we first collected a large set of normal X-rays from both public datasets and our partner hospitals (see Table~\ref{table_dataset} for details). 
Considering the model's generative capabilities, we adopted the currently powerful Diffusion Transformer (DiT) model as our generator and trained it on this collected dataset.

\subsubsection{Lung Lesion Classifier Training on Abnormal X-rays}
With the pre-trained normal X-ray generator established, the subsequent step requires localizing the head-class lesions to guide the inpainting process. To achieve this, we first train a lung lesion classifier on the collected abnormal dataset. We then employ Gradient-weighted Class Activation Mapping (Grad-CAM~\cite {selvaraju2017grad}) to extract the precise regions of interest corresponding to the head classes from this classifier. As illustrated in Figure~\ref{fig_network}, this pre-trained classifier and the associated Grad-CAM module constitute the second step of our pipeline, which is dedicated to lesion area extraction.

\subsection{Tail-class Lesion Data Generation}

The long-tail distribution of the dataset (Figure~\ref{fig_distribute}(a)) leads classifiers to prioritize head-class lesions due to their higher prevalence, consequently resulting in suboptimal performance for the infrequent tail classes. 
Therefore, our objective is to enhance the representation of tail-class samples during training without reducing the number of head-class samples, thereby improving tail-class performance. 
Our core approach is to leverage a pre-trained DiT model to generate new chest X-rays that contain only tail-class lesions. This is achieved by inpainting the head-class lesion regions with normal lung textures, effectively isolating the tail-class lesions. 
Specifically, we utilize the pre-trained classifier $\mathbf{C}$ to generate the class activation maps that localize the head-class lesion areas for inpainting. 
Considering that the classifier has better classification performance for the head classes and the lesion area it generates is more accurate, we extract the CAM maps of the head classes instead of the tail classes:
\begin{equation}
\begin{aligned}
        \vm_{I} &= \textbf{GradCAM}(\mathbf{C}, \mathbf{I}, \mathbf{c}),
\end{aligned}
\end{equation}
where $\mathbf{I}\in \mathbb{R}^{H\times W\times C}$ is the input chest X-ray image; $\vc$ denotes the head class; $\vm_{I}\in \mathbb{R}^{H\times W}$ represents the the lesion activation area of $\vc$ in $\mathbf{I}$.  
After obtaining these CAM maps, we feed them into the pretrained DiT model $\textbf{G}$ to inpaint the lesion area of $\vc$:
\begin{equation}
\begin{aligned}
    \mathbf{I}^{'} &= \textbf{DiT}(\vm_{I}, \mathbf{I}, \mathbf{g}),
\end{aligned}
\end{equation}
where $\mathbf{g}\in \mathbb{R}^{H\times W}$ represents the random Gaussian noise. $\mathbf{I}^{'}\in \mathbb{R}^{H\times W\times C}$ denotes the newly generated image where the head-class lesions have been removed. 
As shown in Figure~\ref{fig_network}, the label of the head class (pink box) will be removed after the inpainting process, and a new training image with the tailed class (green box) is generated.

\subsubsection{LLM Knowledge Guidance}
X-ray is a 2D imaging modality, where 3D anatomies and lesions are projected onto a 2D plane.
Consequently, lesions from different diseases may overlap spatially. This overlap directly impacts our inpainting process: when head-class and tail-class lesions are intertwined within the same region, inpainting the head-class lesion area could inadvertently restore the overlapping tail-class lesions to a normal appearance, thereby generating erroneous training data. Figure~\ref{fig_distribute} visualizes the disease annotations from the VinDr-CXR dataset, which includes 14 disease types, each annotated with a bounding box (bbox). Each point in the figure represents the center of a bbox, The plot overlays lesions from all training samples. Of course, there will be some overlap among lesions. It does not show that on a single image, lesions overlap each other.

To mitigate this issue, we designed a Large Language Model Knowledge Guidance (LKG) module. This module leverages the extensive biomedical knowledge embedded within a large language model (LLM) to identify co-occurrence patterns between head-class and tail-class lesion categories in the current data sample. Specifically, we employ GPT-4 as the underlying LLM for the LKG module. As shown in Figure~\ref{fig_network}, when both head (e.g., red and green boxes) and tail (e.g., pink box) lesion categories co-exist in the labels, the LKG module selectively retains the head class that is most likely to be entangled with the tail-class lesion (green box) while removing other head classes (red box). This strategy effectively prevents the inadvertent erasure of tail-class lesions during the inpainting process.

\begin{figure}
\centering
\includegraphics[width=1.0\linewidth]{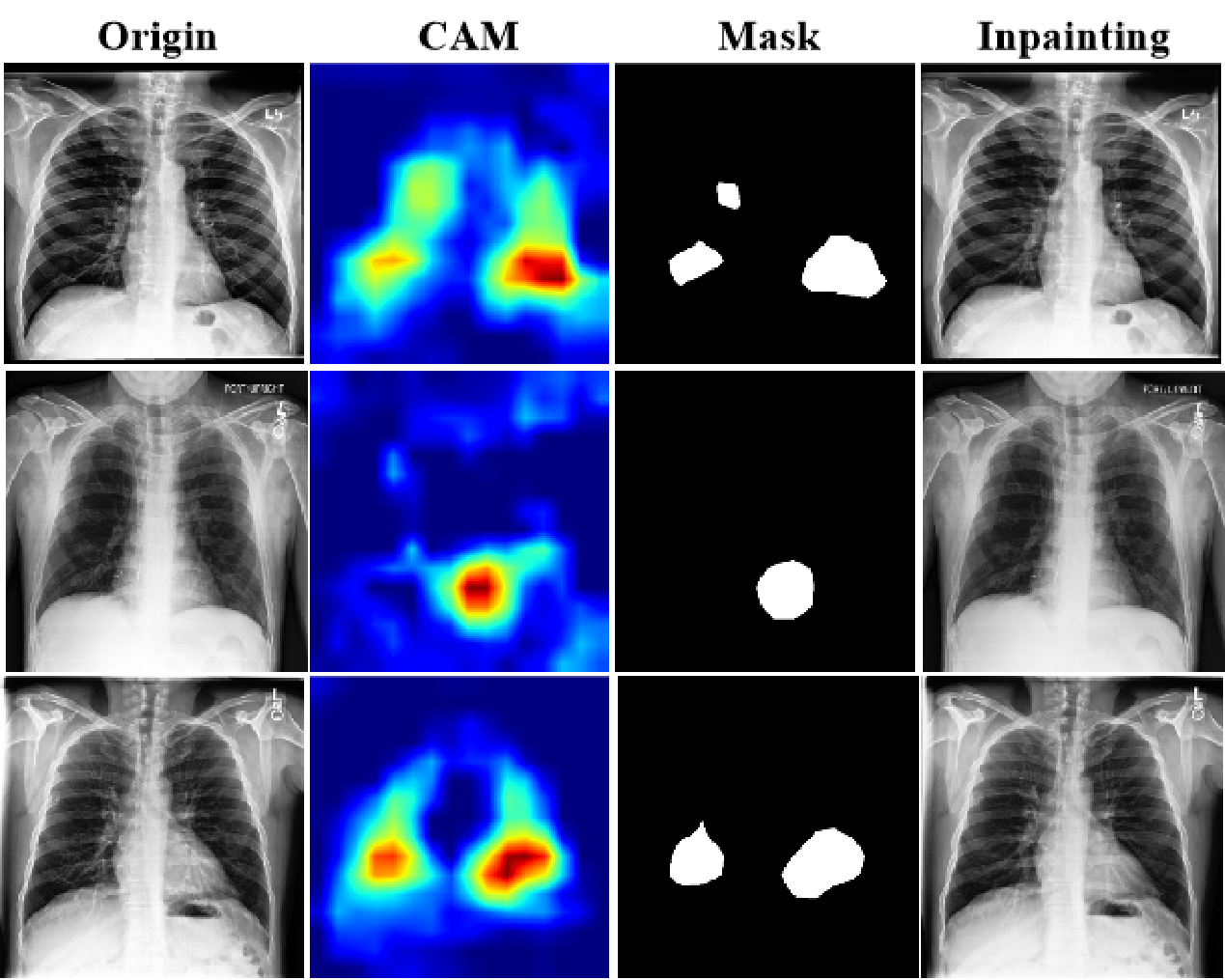}
\caption{Visualization of the inpainting result generated by the normal X-ray diffusion model.} \label{fig_inpating}
\end{figure}

\subsection{Progressive Incremental Finetuning}
As $\textbf{G}$ is trained on a large number of diverse lung samples, the domain of the inpainted images may differ from that of the current training set. When these newly generated images are added to the network for training, catastrophic forgetting will occur, leading to performance degradation in some head classes. 
To address this issue, we designed a simple yet effective training strategy called Progressive Incremental Learning (PIL). 
The original training set and the dataset generated by inpainting are denoted as $\mD_o \in \{\mI_1, \mI_2, ..., \mI_N\}$ and $\mD_i\in \{\mI^{'}_1, \mI^{'}_2, ..., \mI^{'}_N\}$. During the initial training phase, $\mD_i$ and $\mD_o$ are combined into the training set at a lower ratio to mitigate the impact of domain differences:
\begin{equation}
\begin{aligned}
    \mD &= \mD_o + \mD_i(1-e^{-\beta n}),
\end{aligned}
\end{equation}
where $\mD$ represents the whole training dataset. $\beta$ is a hyperparameter that controls the magnitude of $\mD_i$; $n$ denotes the number of training epoch. As the number of training epochs increases, more and more $\mD_i$ are added to the training set. Through this progressive training method, the network can gradually improve the performance of the tail class while retaining the performance of the head classes.


\section{Experiments}
\subsection{Datasets}
\textbf{Normal X-ray Dataset.} 
We extract normal X-rays from the public datasets~\cite{nguyen2022vindr,irvin2019chexpert,wang2017chestx,johnson2019mimic,bustos2020padchest}, to form a rich, multi-source normal X-ray dataset. 
Specifically, 215,652 normal X-ray images were collected from 7 publicly available datasets.
In addition, we also collect a large number of normal X-rays (105,234) from our partner hospitals. The dataset details are given in Table~\ref{table_dataset}.

\begin{table}[]
\caption{The number of normal samples in the collected chest X-ray dataset.}\label{table_dataset}
\centering
\scalebox{0.9}{ 
\begin{tabular}{c|c|c}
\toprule
Dataset                                             & Public        & Number of Normal Samples \\ \hline
Pneumonia Classification Challenge                    & \Checkmark       & 1584     \\ \hline
RSNA Pneumonia                                        & \Checkmark       & 8852     \\ \hline
VinDr-CXR~\cite{nguyen2022vindr}                      & \Checkmark       & 10606     \\ \hline
CheXpert~\cite{irvin2019chexpert}                     & \Checkmark       & 17002    \\ \hline
NIH-CXR~\cite{wang2017chestx}                         & \Checkmark       & 60362    \\ \hline
MIMIC-CXR~\cite{johnson2019mimic}                     & \Checkmark       & 81117    \\ \hline
PadChest~\cite{bustos2020padchest}                    & \Checkmark       & 36129         \\ \hline
Private Data                            & \ding{55}        & 105434   \\ \hline
Total Number                                          &                  & 321086   \\ 
\bottomrule
\end{tabular}
}
\end{table}

\noindent \textbf{Evaluation Dataset.} We use two public chest X-ray datasets (MIMIC-CXR\cite{johnson2019mimic} and CheXpert~\cite{irvin2019chexpert}) to evaluate the performance of our proposed method. Both datasets have the same 13 lesion categories and consistent long-tail distribution, as shown in Figure~\ref{fig_distribute}(a). The 13 lesion categories are Enlarged Cardiomediastinum (EC), Cardiomegaly (CA), Lung Opacity (LO), Lung Lesion (LL), Edema (ED), Consolidation (CO), Pneumonia (PA), Atelectasis (AT), Pneumothorax (PX), Pleural Effusion (PE), Pleural Other (PO), Fracture (FE), and Support Devices (SS). 
TThere are 150,904 samples in MIMIC-CXR and 166,458 samples in CheXpert.
We randomly selected 80\% of the dataset for training sets and the remaining 20\% for testing.

\subsection{Implementation Details}
PyTorch is used for training and testing. For the training of classification network, we use a batch size of 128, Adam optimizer and a learning rate of 0.001 for training. The total number of training epochs and fine-tuning epochs is set as 10. All input images are resized to 256$\times$256 pixels, and no additional data augmentation methods are uesed. 
The cross-entropy loss function is used for training all classification networks. For the training of DiT, we resize all the image to 512$\times$512 pixels and the model is DiT-XL/2. The training configurations are consistent with the official code. All experiments are conducted on an NVIDIA V100 GPU. 

\begin{table*}[]
\caption{Comparison of the performance of different classification networks using our proposed data augmentation method on the MIMIC-CXR 
 and CheXpert datasets. The categories marked in red are the tail classes. 'Aug' denotes the proposed normal X-ray inpainting augmentation.}\label{table_mimic} 
 \centering
\scalebox{0.38}{
\begin{tabular}{c|c|c|ccccccccccccc|c}
\toprule
                             &                                   &                       & \multicolumn{13}{c|}{Disease Category}                                                                                                                                                                                                                                                                                                                                                                                                                                                                                                                                              &                                          \\ \cline{4-16}
\multirow{-2}{*}{Dataset}    & \multirow{-2}{*}{Network}         & \multirow{-2}{*}{Aug} & \multicolumn{1}{c|}{{\color[HTML]{FE0000} EC}}                & \multicolumn{1}{c|}{CA}    & \multicolumn{1}{c|}{LO}    & \multicolumn{1}{c|}{{\color[HTML]{FE0000} LL}}                & \multicolumn{1}{c|}{ED}    & \multicolumn{1}{c|}{{\color[HTML]{FE0000} CO}}                & \multicolumn{1}{c|}{{\color[HTML]{FE0000} PA}}                & \multicolumn{1}{c|}{AT}    & \multicolumn{1}{c|}{PX}    & \multicolumn{1}{c|}{PE}    & \multicolumn{1}{c|}{{\color[HTML]{FE0000} PO}}                & \multicolumn{1}{c|}{{\color[HTML]{FE0000} FE}}                & SS    & \multirow{-2}{*}{F1 Score}               \\ \hline
                             &                                   &                       & \multicolumn{1}{c|}{7.22}                                     & \multicolumn{1}{c|}{43.15} & \multicolumn{1}{c|}{69.73} & \multicolumn{1}{c|}{10.34}                                     & \multicolumn{1}{c|}{45.8} & \multicolumn{1}{c|}{4.2}                                     & \multicolumn{1}{c|}{0.21}                                     & \multicolumn{1}{c|}{19.72} & \multicolumn{1}{c|}{44.81} & \multicolumn{1}{c|}{73.31} & \multicolumn{1}{c|}{4.02}                                     & \multicolumn{1}{c|}{11.9}                                    & 78.06 & 31.72                                    \\ \cline{3-17} 
                             & \multirow{-2}{*}{ResNet-50}       & \Checkmark                 & \multicolumn{1}{c|}{12.88(+5.66)}  & \multicolumn{1}{c|}{38.7} & \multicolumn{1}{c|}{71.32} & \multicolumn{1}{c|}{18.01(+7.67)} & \multicolumn{1}{c|}{54.14} & \multicolumn{1}{c|}{13.54(+9.34)} & \multicolumn{1}{c|}{9.31(+9.1)}  & \multicolumn{1}{c|}{31.21} & \multicolumn{1}{c|}{39.7} & \multicolumn{1}{c|}{71.68} & \multicolumn{1}{c|}{9.03(+5.01)} & \multicolumn{1}{c|}{18.62(+6.72)} & 74.83 & 35.61(+3.89) \\ \cline{2-17} 
                             &                                   &                       & \multicolumn{1}{c|}{17.05}                                         & \multicolumn{1}{c|}{47.15}      & \multicolumn{1}{c|}{72.64}      & \multicolumn{1}{c|}{21.36}                                         & \multicolumn{1}{c|}{57.72}      & \multicolumn{1}{c|}{20.96}                                         & \multicolumn{1}{c|}{13.09}                                         & \multicolumn{1}{c|}{34.66}      & \multicolumn{1}{c|}{45.07}      & \multicolumn{1}{c|}{70.71}      & \multicolumn{1}{c|}{13.74}                                         & \multicolumn{1}{c|}{22.97}                                         & 78.83      & 39.68                                         \\ \cline{3-17} 
                             & \multirow{-2}{*}{EfficientNet-B0} & \Checkmark                 & \multicolumn{1}{c|}{18.77(+1.72)}                                         & \multicolumn{1}{c|}{48.62}      & \multicolumn{1}{c|}{72.36}      & \multicolumn{1}{c|}{22.75(+1.39)}                                         & \multicolumn{1}{c|}{59.66}      & \multicolumn{1}{c|}{22.11(+1.15)}                                         & \multicolumn{1}{c|}{13.91(+0.82)}                                         & \multicolumn{1}{c|}{35.26}      & \multicolumn{1}{c|}{47.04}      & \multicolumn{1}{c|}{70
                             .39}      & \multicolumn{1}{c|}{14.78(+1.04)}                                         & \multicolumn{1}{c|}{25.04(+2.07)}                                         &  80.82     & 40.88(+1.2)                                         \\ \cline{2-17} 
                             &                                   &                       & \multicolumn{1}{c|}{16.19}                                    & \multicolumn{1}{c|}{46.82} & \multicolumn{1}{c|}{72.27} & \multicolumn{1}{c|}{18.89}                                    & \multicolumn{1}{c|}{53.89} & \multicolumn{1}{c|}{19.17}                                    & \multicolumn{1}{c|}{11.51}                                    & \multicolumn{1}{c|}{33.44} & \multicolumn{1}{c|}{38.61} & \multicolumn{1}{c|}{68.07} & \multicolumn{1}{c|}{10.21}                                    & \multicolumn{1}{c|}{19.61}                                    & 76.78 &  37.34                                   \\ \cline{3-17} 
                             & \multirow{-2}{*}{ConvNext-B}   & \Checkmark                 & \multicolumn{1}{c|}{19.89(+3.7)}  & \multicolumn{1}{c|}{49.62} & \multicolumn{1}{c|}{72.39} & \multicolumn{1}{c|}{21.69(+2.8)}  & \multicolumn{1}{c|}{55.52} & \multicolumn{1}{c|}{21.43(+2.26)} & \multicolumn{1}{c|}{13.56(+2.05)} & \multicolumn{1}{c|}{35.42} & \multicolumn{1}{c|}{39.32} & \multicolumn{1}{c|}{71.23} & \multicolumn{1}{c|}{13.69(+3.48)} & \multicolumn{1}{c|}{24.25(+4.64)} & 75.04 & 39.46(+2.12) \\ \cline{2-17} 
                             &                                   &                       & \multicolumn{1}{c|}{13.67}                                         & \multicolumn{1}{c|}{48.18}      & \multicolumn{1}{c|}{72.27}      & \multicolumn{1}{c|}{15.42}                                         & \multicolumn{1}{c|}{54.06}      & \multicolumn{1}{c|}{10.92}                                         & \multicolumn{1}{c|}{0.41}                                         & \multicolumn{1}{c|}{32.34}      & \multicolumn{1}{c|}{35.16}      & \multicolumn{1}{c|}{66.17}      & \multicolumn{1}{c|}{0}                                         & \multicolumn{1}{c|}{19.53}                                         & 74.86      &  34.07                                        \\ \cline{3-17} 
                             & \multirow{-2}{*}{Swin-S}       & \Checkmark                 & \multicolumn{1}{c|}{19.21(+5.54)}                                         & \multicolumn{1}{c|}{48.74}      & \multicolumn{1}{c|}{72.27}      & \multicolumn{1}{c|}{20.83(+5.41)}                                         & \multicolumn{1}{c|}{56.01}      & \multicolumn{1}{c|}{20.97(+10.05)}                                         & \multicolumn{1}{c|}{13.81(+13.4)}                                         & \multicolumn{1}{c|}{34.09}      & \multicolumn{1}{c|}{42.62}      & \multicolumn{1}{c|}{70.76}      & \multicolumn{1}{c|}{12.81(+12.81)}                                         & \multicolumn{1}{c|}{24.75(+5.22)}                                         & 76.53      & 39.49(+5.42)                                         \\ \cline{2-17} 
                             &                                   &                       & \multicolumn{1}{c|}{2.96}                                     & \multicolumn{1}{c|}{36.25} & \multicolumn{1}{c|}{72.27} & \multicolumn{1}{c|}{11.16}                                    & \multicolumn{1}{c|}{51.66} & \multicolumn{1}{c|}{0.14}                                     & \multicolumn{1}{c|}{0}                                     & \multicolumn{1}{c|}{32.09} & \multicolumn{1}{c|}{29.43} & \multicolumn{1}{c|}{63.84} & \multicolumn{1}{c|}{0.71}                                     & \multicolumn{1}{c|}{12.27}                                    & 74.86 & 29.81                                    \\ \cline{3-17} 
\multirow{-10}{*}{CheXpert} & \multirow{-2}{*}{ViT-B}        & \Checkmark                 & \multicolumn{1}{c|}{11.72(+8.76)} & \multicolumn{1}{c|}{41.23} & \multicolumn{1}{c|}{72.27} & \multicolumn{1}{c|}{16.32(+5.16)} & \multicolumn{1}{c|}{52.91} & \multicolumn{1}{c|}{9.23(+9.09)} & \multicolumn{1}{c|}{5.64(+5.64)}  & \multicolumn{1}{c|}{32.92} & \multicolumn{1}{c|}{31.14} & \multicolumn{1}{c|}{66.39} & \multicolumn{1}{c|}{9.27(+8.56)}  & \multicolumn{1}{c|}{18.29(+6.02)} & 74.88 & 34.01(+4.2) \\ \hline
                             &                                   &                       & \multicolumn{1}{c|}{4.69}                                         & \multicolumn{1}{c|}{47.61}      & \multicolumn{1}{c|}{51.17}      & \multicolumn{1}{c|}{9.28}                                         & \multicolumn{1}{c|}{38.52}      & \multicolumn{1}{c|}{1.41}                                         & \multicolumn{1}{c|}{14.53}                                         & \multicolumn{1}{c|}{38.89}      & \multicolumn{1}{c|}{43.45}      & \multicolumn{1}{c|}{63.41}      & \multicolumn{1}{c|}{3.81}                                         & \multicolumn{1}{c|}{12.03}                                         & 75.57      & 31.11                                         \\ \cline{3-17} 
                             & \multirow{-2}{*}{ResNet-50}       & \Checkmark                 & \multicolumn{1}{c|}{13.06(+8.37)}                                         & \multicolumn{1}{c|}{47.69}      & \multicolumn{1}{c|}{50.42}      & \multicolumn{1}{c|}{14.77(+5.49)}                                         & \multicolumn{1}{c|}{40.53}      & \multicolumn{1}{c|}{12.48(+11.07)}                                         & \multicolumn{1}{c|}{17.19(+2.66)}                                         & \multicolumn{1}{c|}{39.41}      & \multicolumn{1}{c|}{43.32}      & \multicolumn{1}{c|}{64.27}      & \multicolumn{1}{c|}{11.35(+7.54)}                                         & \multicolumn{1}{c|}{14.16(+2.13)}                                         & 72.24      & 33.91(+2.8)                                         \\ \cline{2-17} 
                             &                                   &                       & \multicolumn{1}{c|}{15.96}                                         & \multicolumn{1}{c|}{52.67}      & \multicolumn{1}{c|}{53.76}      & \multicolumn{1}{c|}{23.12}                                         & \multicolumn{1}{c|}{49.95}      & \multicolumn{1}{c|}{21.61}                                         & \multicolumn{1}{c|}{25.97}                                         & \multicolumn{1}{c|}{52.66}      & \multicolumn{1}{c|}{47.17}      & \multicolumn{1}{c|}{66.52}      & \multicolumn{1}{c|}{13.17}                                         & \multicolumn{1}{c|}{21.01}                                         & 74.23      & 40.04                                         \\ \cline{3-17} 
                             & \multirow{-2}{*}{EfficientNet-B0} & \Checkmark                 & \multicolumn{1}{c|}{16.62(+0.66)}                                         & \multicolumn{1}{c|}{52.74}      & \multicolumn{1}{c|}{53.71}      & \multicolumn{1}{c|}{25.52(+2.4)}                                         & \multicolumn{1}{c|}{49.11}      & \multicolumn{1}{c|}{23.73(+2.12)}                                         & \multicolumn{1}{c|}{26.62(+0.65)}                                         & \multicolumn{1}{c|}{52.71}      & \multicolumn{1}{c|}{47.75}      & \multicolumn{1}{c|}{67.84}      & \multicolumn{1}{c|}{15.74(+2.57)}                                         & \multicolumn{1}{c|}{22.22(+1.21)}                                         & 72.31      & 40.51(+0.47)                                         \\ \cline{2-17} 
                             &                                   &                       & \multicolumn{1}{c|}{13.58}                                         & \multicolumn{1}{c|}{53.34}      & \multicolumn{1}{c|}{53.31}      & \multicolumn{1}{c|}{18.94}                                         & \multicolumn{1}{c|}{47.08}      & \multicolumn{1}{c|}{21.12}                                         & \multicolumn{1}{c|}{24.59}                                         & \multicolumn{1}{c|}{52.87}      & \multicolumn{1}{c|}{34.93}      & \multicolumn{1}{c|}{64.93}      & \multicolumn{1}{c|}{9.64}                                         & \multicolumn{1}{c|}{16.16}                                         & 66.25      &  36.68                                        \\ \cline{3-17} 
                             & \multirow{-2}{*}{ConvNext-B}   & \Checkmark                 & \multicolumn{1}{c|}{16.35(+2.77)}                                         & \multicolumn{1}{c|}{53.61}      & \multicolumn{1}{c|}{53.64}      & \multicolumn{1}{c|}{21.03(+2.09)}                                         & \multicolumn{1}{c|}{49.89}      & \multicolumn{1}{c|}{20.34(+0.78)}                                         & \multicolumn{1}{c|}{25.16(+0.57)}                                         & \multicolumn{1}{c|}{52.03}      & \multicolumn{1}{c|}{39.54}      & \multicolumn{1}{c|}{66.2}      & \multicolumn{1}{c|}{11.12(+1.48)}                                         & \multicolumn{1}{c|}{17.79(+1.63)}                                         &  68.71     &  38.11(+1.43)                                        \\ \cline{2-17} 
                             &                                   &                       & \multicolumn{1}{c|}{0.84}                                         & \multicolumn{1}{c|}{50.63}      & \multicolumn{1}{c|}{53.28}      & \multicolumn{1}{c|}{16.09}                                         & \multicolumn{1}{c|}{45.57}      & \multicolumn{1}{c|}{6.09}                                         & \multicolumn{1}{c|}{20.41}                                         & \multicolumn{1}{c|}{49.8}      & \multicolumn{1}{c|}{23.51}      & \multicolumn{1}{c|}{58.16}      & \multicolumn{1}{c|}{0.48}                                         & \multicolumn{1}{c|}{12.13}                                         & 61.04      & 30.61                                         \\ \cline{3-17} 
                             & \multirow{-2}{*}{Swin-S}       & \Checkmark                 & \multicolumn{1}{c|}{13.92(+13.08)}                                         & \multicolumn{1}{c|}{53.79}      & \multicolumn{1}{c|}{53.38}      & \multicolumn{1}{c|}{19.37(+3.28)}                                         & \multicolumn{1}{c|}{49.73}      & \multicolumn{1}{c|}{20.81(+14.72)}                                         & \multicolumn{1}{c|}{24.98(+4.57)}                                             & \multicolumn{1}{c|}{51.54}      & \multicolumn{1}{c|}{28.83}      & \multicolumn{1}{c|}{65.32}      & \multicolumn{1}{c|}{8.57(+8.09)}                                         & \multicolumn{1}{c|}{16.12(+3.99)}                                         & 64.65      &  36.23(+5.62)                                        \\ \cline{2-17} 
                             &                                   &                       & \multicolumn{1}{c|}{0.12}                                         & \multicolumn{1}{c|}{50.52}      & \multicolumn{1}{c|}{53.28}      & \multicolumn{1}{c|}{12.22}                                         & \multicolumn{1}{c|}{45.58}      & \multicolumn{1}{c|}{0.08}                                         & \multicolumn{1}{c|}{20.91}                                         & \multicolumn{1}{c|}{48.93}      & \multicolumn{1}{c|}{14.79}      & \multicolumn{1}{c|}{58.4}      & \multicolumn{1}{c|}{0}                                         & \multicolumn{1}{c|}{9.25}                                         & 60.98      & 28.85                                         \\ \cline{3-17} 
\multirow{-10}{*}{MIMIC}     & \multirow{-2}{*}{ViT-B}        & \Checkmark                 & \multicolumn{1}{c|}{7.34(+7.22)}                                         & \multicolumn{1}{c|}{51.73}      & \multicolumn{1}{c|}{53.76}      & \multicolumn{1}{c|}{16.73(+13.08)}                                        & \multicolumn{1}{c|}{44.53}      & \multicolumn{1}{c|}{10.94(+10.86)}                                         & \multicolumn{1}{c|}{22.05(+1.14)}                                         & \multicolumn{1}{c|}{49.61}      & \multicolumn{1}{c|}{21.65}      & \multicolumn{1}{c|}{58.76}      & \multicolumn{1}{c|}{3.91(+3.91)}                                         & \multicolumn{1}{c|}{12.16(+2.91)}                                         & 62.51      & 31.97(+3.12)                                         \\ \bottomrule
\end{tabular}
 }
\end{table*}

\begin{table*}[]
\caption{Performance comparison of data augmentation methods based on captioning and inpainting. The categories marked in red are the tail classes. }\label{table_caption} 
\centering
\scalebox{0.37}{
\begin{tabular}{c|c|c|ccccccccccccc|c}
\toprule
                           &                           &                          & \multicolumn{13}{c|}{Disease Category}                                                                                                                                                                                                                                                                                                                                                                                                                                                    &                            \\ \cline{4-16}
\multirow{-2}{*}{Dataset}  & \multirow{-2}{*}{Network} & \multirow{-2}{*}{Method} & \multicolumn{1}{c|}{{\color[HTML]{FE0000} EC}} & \multicolumn{1}{c|}{CA}    & \multicolumn{1}{c|}{LO}    & \multicolumn{1}{c|}{{\color[HTML]{FE0000} LL}} & \multicolumn{1}{c|}{ED}    & \multicolumn{1}{c|}{{\color[HTML]{FE0000} CO}} & \multicolumn{1}{c|}{{\color[HTML]{FE0000} PA}} & \multicolumn{1}{c|}{AT}    & \multicolumn{1}{c|}{PX}    & \multicolumn{1}{c|}{PE}    & \multicolumn{1}{c|}{{\color[HTML]{FE0000} PO}} & \multicolumn{1}{c|}{{\color[HTML]{FE0000} FE}} & SS    & \multirow{-2}{*}{F1 Score} \\ \hline
                           &                           & -                        & \multicolumn{1}{c|}{7.22}                      & \multicolumn{1}{c|}{43.15} & \multicolumn{1}{c|}{69.73} & \multicolumn{1}{c|}{10.34}                     & \multicolumn{1}{c|}{45.8}  & \multicolumn{1}{c|}{4.2}                       & \multicolumn{1}{c|}{0.21}                      & \multicolumn{1}{c|}{19.72} & \multicolumn{1}{c|}{44.81} & \multicolumn{1}{c|}{73.31} & \multicolumn{1}{c|}{4.02}                      & \multicolumn{1}{c|}{11.9}                      & 78.06 & 31.72                      \\ \cline{3-17} 
                           &                           & DiT-Caption              & \multicolumn{1}{c|}{8.53(+1.31)}               & \multicolumn{1}{c|}{36.24} & \multicolumn{1}{c|}{70.88} & \multicolumn{1}{c|}{12.52(+2.18)}              & \multicolumn{1}{c|}{53.16} & \multicolumn{1}{c|}{6.27(+2.07)}               & \multicolumn{1}{c|}{1.3(+1.09)}                & \multicolumn{1}{c|}{20.57} & \multicolumn{1}{c|}{44.39} & \multicolumn{1}{c|}{72.56} & \multicolumn{1}{c|}{3.76(-0.26)}               & \multicolumn{1}{c|}{11.62(-0.28)}              & 75.31 & 32.08(+0.36)               \\ \cline{3-17} 
\multirow{-3}{*}{CheXpert} &                           & DiT-Inpainting(our)      & \multicolumn{1}{c|}{12.88(+5.66)}              & \multicolumn{1}{c|}{38.7}  & \multicolumn{1}{c|}{71.32} & \multicolumn{1}{c|}{18.01(+7.67)}              & \multicolumn{1}{c|}{54.14} & \multicolumn{1}{c|}{13.54(+9.34)}              & \multicolumn{1}{c|}{9.31(+9.1)}                & \multicolumn{1}{c|}{31.21} & \multicolumn{1}{c|}{39.7}  & \multicolumn{1}{c|}{71.68} & \multicolumn{1}{c|}{9.03(+5.01)}               & \multicolumn{1}{c|}{18.62(+6.72)}              & 74.83 & 35.61(+3.89)               \\ \cline{1-1} \cline{3-17} 
                           &                           & -                        & \multicolumn{1}{c|}{4.69}                      & \multicolumn{1}{c|}{47.61} & \multicolumn{1}{c|}{51.17} & \multicolumn{1}{c|}{9.28}                      & \multicolumn{1}{c|}{38.52} & \multicolumn{1}{c|}{1.41}                      & \multicolumn{1}{c|}{14.53}                     & \multicolumn{1}{c|}{38.89} & \multicolumn{1}{c|}{43.45} & \multicolumn{1}{c|}{63.41} & \multicolumn{1}{c|}{3.81}                      & \multicolumn{1}{c|}{12.03}                     & 75.57 & 31.11                      \\ \cline{3-17} 
                           &                           & DiT-Caption              & \multicolumn{1}{c|}{6.32(+1.63)}               & \multicolumn{1}{c|}{46.58} & \multicolumn{1}{c|}{50.87} & \multicolumn{1}{c|}{10.32(+1.04)}              & \multicolumn{1}{c|}{38.6}  & \multicolumn{1}{c|}{3.52(+2.11)}               & \multicolumn{1}{c|}{13.57(-0.96)}              & \multicolumn{1}{c|}{40.3}  & \multicolumn{1}{c|}{44.87} & \multicolumn{1}{c|}{62.16} & \multicolumn{1}{c|}{4.02(+0.21)}               & \multicolumn{1}{c|}{12.85(+0.82)}              & 74.34 & 31.4(+0.29)                \\ \cline{3-17} 
\multirow{-3}{*}{MIMIC}    & \multirow{-6}{*}{R50}     & DiT-Inpainting(our)      & \multicolumn{1}{c|}{13.06(+8.37)}              & \multicolumn{1}{c|}{47.69} & \multicolumn{1}{c|}{50.42} & \multicolumn{1}{c|}{14.77(+5.49)}              & \multicolumn{1}{c|}{40.53} & \multicolumn{1}{c|}{12.48(+11.07)}             & \multicolumn{1}{c|}{17.19(2.66)}               & \multicolumn{1}{c|}{39.41} & \multicolumn{1}{c|}{43.32} & \multicolumn{1}{c|}{64.27} & \multicolumn{1}{c|}{11.35(+7.54)}              & \multicolumn{1}{c|}{14.16(+2.13)}              & 72.24 & 33.91(+2.8)                \\ \bottomrule
\end{tabular}
 }
\end{table*}

\subsection{Performance Analysis}
\textbf{Improvement in Different Classification Networks.} 
To evaluate the effectiveness of our proposed normal data inpainting method, we apply this data augmentation on different classification networks (i.e., CNN-based networks, including ResNet~\cite{he2016deep}, EfficientNet~\cite{tan2019efficientnet}, Convnext~\cite{liu2022convnet} and Transformer-based networks, Swin Transformer~\cite{liu2021swin} and ViT~\cite{dosovitskiy2020image}), results are given in Table~\ref{table_mimic}. 
From the table we can see that the performance of CNN-based networks is better than that of transformer-based networks, where EfficientNet shows the best performance on both datasets, 40.88\% and 40.51\%, respectively, while ViT has the worst performance, 34.01\% and 31.97\%, respectively. 
In terms of the performance improvement of the tail class, after the inpainting tail classes data was added to the network training, the performance of the tail classes was greatly improved. 
Although the performance of the head classes has variations, it is clinically acceptable to lose some of the performance of the head classes in exchange for a significant improvement in the performance of the tail classes. These results proved the effectiveness of our proposed data augmentation method of tail classes.
\\

\begin{table*}[]
\caption{Ablation studies of different loss functions under the proposed progressive incremental learning strategy. 'Aug' denotes the proposed normal X-ray inpainting augmentation.}\label{table_loss_ablation}
\centering
\scalebox{0.47}{
\begin{tabular}{c|c|c|c|c|c|ccccccccccccc|c}
\toprule
                             &                            &                      &                      &                       &                       & \multicolumn{13}{c|}{Disease Category}                                                                                                                                                                                                                                                                                                                                                                                                                                                                                                                                                                                                &                            \\ \cline{7-19}
\multirow{-2}{*}{Network}    & \multirow{-2}{*}{Dataset}  & \multirow{-2}{*}{$\mathcal{L}_c$} & \multirow{-2}{*}{$\mathcal{L}_f$} & \multirow{-2}{*}{Aug} & \multirow{-2}{*}{PIL} & \multicolumn{1}{c|}{{\color[HTML]{FE0000} EC}} & \multicolumn{1}{c|}{{\color[HTML]{000000} CA}} & \multicolumn{1}{c|}{{\color[HTML]{000000} LO}} & \multicolumn{1}{c|}{{\color[HTML]{FE0000} LL}} & \multicolumn{1}{c|}{{\color[HTML]{000000} ED}} & \multicolumn{1}{c|}{{\color[HTML]{FE0000} CO}} & \multicolumn{1}{c|}{{\color[HTML]{FE0000} PA}} & \multicolumn{1}{c|}{{\color[HTML]{000000} AT}} & \multicolumn{1}{c|}{{\color[HTML]{000000} PX}} & \multicolumn{1}{c|}{{\color[HTML]{000000} PE}} & \multicolumn{1}{c|}{{\color[HTML]{FE0000} PO}} & \multicolumn{1}{c|}{{\color[HTML]{FE0000} FE}} & {\color[HTML]{000000} SS} & \multirow{-2}{*}{F1 Score} \\ \hline
                             &                            & \Checkmark                  &                      &                       &                       & \multicolumn{1}{c|}{7.22}                      & \multicolumn{1}{c|}{43.15}                     & \multicolumn{1}{c|}{69.73}                     & \multicolumn{1}{c|}{10.34}                     & \multicolumn{1}{c|}{45.8}                      & \multicolumn{1}{c|}{4.2}                       & \multicolumn{1}{c|}{0.21}                      & \multicolumn{1}{c|}{19.72}                     & \multicolumn{1}{c|}{44.81}                     & \multicolumn{1}{c|}{73.31}                     & \multicolumn{1}{c|}{4.02}                      & \multicolumn{1}{c|}{11.9}                      & 78.06                     & 31.72                      \\ \cline{3-20} 
                             &                            & \Checkmark                  &                      & \Checkmark                   &                       & \multicolumn{1}{c|}{10.31}                     & \multicolumn{1}{c|}{36.19}                     & \multicolumn{1}{c|}{64.94}                     & \multicolumn{1}{c|}{14.51}                     & \multicolumn{1}{c|}{40.28}                     & \multicolumn{1}{c|}{8.92}                      & \multicolumn{1}{c|}{8.27}                      & \multicolumn{1}{c|}{18.68}                     & \multicolumn{1}{c|}{38.49}                     & \multicolumn{1}{c|}{67.72}                     & \multicolumn{1}{c|}{6.34}                      & \multicolumn{1}{c|}{14.66}                     & 73.22                     & 30.96                      \\ \cline{3-20} 
                             &                            & \Checkmark                  &                      & \Checkmark                   & \Checkmark                   & \multicolumn{1}{c|}{12.88}                     & \multicolumn{1}{c|}{38.7}                      & \multicolumn{1}{c|}{71.32}                     & \multicolumn{1}{c|}{18.01}                     & \multicolumn{1}{c|}{54.14}                     & \multicolumn{1}{c|}{13.54}                     & \multicolumn{1}{c|}{9.31}                      & \multicolumn{1}{c|}{31.21}                     & \multicolumn{1}{c|}{39.7}                      & \multicolumn{1}{c|}{71.68}                     & \multicolumn{1}{c|}{9.03}                      & \multicolumn{1}{c|}{18.62}                     & 74.83                     & 35.61                      \\ \cline{3-20} 
                             &                            &                      & \Checkmark                  &                       &                       & \multicolumn{1}{c|}{6.33}                      & \multicolumn{1}{c|}{42.49}                     & \multicolumn{1}{c|}{58.31}                     & \multicolumn{1}{c|}{4.21}                      & \multicolumn{1}{c|}{54.52}                     & \multicolumn{1}{c|}{5.18}                      & \multicolumn{1}{c|}{0.79}                      & \multicolumn{1}{c|}{11.28}                     & \multicolumn{1}{c|}{46.87}                     & \multicolumn{1}{c|}{69.26}                     & \multicolumn{1}{c|}{5.31}                      & \multicolumn{1}{c|}{13.43}                     & 79.84                     & 30.6                       \\ \cline{3-20} 
                             &                            &                      & \Checkmark                  & \Checkmark                   &                       & \multicolumn{1}{c|}{9.24}                      & \multicolumn{1}{c|}{38.93}                     & \multicolumn{1}{c|}{53.18}                     & \multicolumn{1}{c|}{14.72}                     & \multicolumn{1}{c|}{51.69}                     & \multicolumn{1}{c|}{8.27}                      & \multicolumn{1}{c|}{9.21}                      & \multicolumn{1}{c|}{12.19}                     & \multicolumn{1}{c|}{41.11}                     & \multicolumn{1}{c|}{62.38}                     & \multicolumn{1}{c|}{8.13}                      & \multicolumn{1}{c|}{15.59}                     & 73.36                     & 30.61                      \\ \cline{3-20} 
                             & \multirow{-6}{*}{CheXpert} &                      & \Checkmark                  & \Checkmark                   & \Checkmark                   & \multicolumn{1}{c|}{11.31}                     & \multicolumn{1}{c|}{43.13}                     & \multicolumn{1}{c|}{57.62}                     & \multicolumn{1}{c|}{16.83}                     & \multicolumn{1}{c|}{56.36}                     & \multicolumn{1}{c|}{9.52}                      & \multicolumn{1}{c|}{9.62}                      & \multicolumn{1}{c|}{12.57}                     & \multicolumn{1}{c|}{48.94}                     & \multicolumn{1}{c|}{68.33}                     & \multicolumn{1}{c|}{9.46}                      & \multicolumn{1}{c|}{16.03}                     & 78.21                     & 33.68                      \\ \cline{2-20} 
                             &                            & \Checkmark                  &                      &                       &                       & \multicolumn{1}{c|}{4.69}                      & \multicolumn{1}{c|}{47.61}                     & \multicolumn{1}{c|}{51.17}                     & \multicolumn{1}{c|}{9.28}                      & \multicolumn{1}{c|}{38.52}                     & \multicolumn{1}{c|}{1.41}                      & \multicolumn{1}{c|}{14.53}                     & \multicolumn{1}{c|}{38.89}                     & \multicolumn{1}{c|}{43.45}                     & \multicolumn{1}{c|}{63.41}                     & \multicolumn{1}{c|}{3.81}                      & \multicolumn{1}{c|}{12.03}                     & 75.57                     & 31.11                      \\ \cline{3-20} 
                             &                            & \Checkmark                  &                      & \Checkmark                   &                       & \multicolumn{1}{c|}{8.14}                      & \multicolumn{1}{c|}{43.29}                     & \multicolumn{1}{c|}{48.81}                     & \multicolumn{1}{c|}{12.34}                     & \multicolumn{1}{c|}{35.69}                     & \multicolumn{1}{c|}{7.62}                      & \multicolumn{1}{c|}{14.8}                      & \multicolumn{1}{c|}{32.11}                     & \multicolumn{1}{c|}{40.18}                     & \multicolumn{1}{c|}{62.19}                     & \multicolumn{1}{c|}{6.93}                      & \multicolumn{1}{c|}{12.26}                     & 69.31                     & 30.28                      \\ \cline{3-20} 
                             &                            & \Checkmark                  &                      & \Checkmark                   & \Checkmark                   & \multicolumn{1}{c|}{13.06}                     & \multicolumn{1}{c|}{47.69}                     & \multicolumn{1}{c|}{50.42}                     & \multicolumn{1}{c|}{14.77}                     & \multicolumn{1}{c|}{40.53}                     & \multicolumn{1}{c|}{12.48}                     & \multicolumn{1}{c|}{17.19}                     & \multicolumn{1}{c|}{39.41}                     & \multicolumn{1}{c|}{43.32}                     & \multicolumn{1}{c|}{64.27}                     & \multicolumn{1}{c|}{11.35}                     & \multicolumn{1}{c|}{14.16}                     & 72.24                     & 33.91                      \\ \cline{3-20} 
                             &                            &                      & \Checkmark                  &                       &                       & \multicolumn{1}{c|}{1.22}                      & \multicolumn{1}{c|}{48.83}                     & \multicolumn{1}{c|}{53.12}                     & \multicolumn{1}{c|}{5.19}                      & \multicolumn{1}{c|}{43.92}                     & \multicolumn{1}{c|}{4.12}                      & \multicolumn{1}{c|}{15.42}                     & \multicolumn{1}{c|}{38.12}                     & \multicolumn{1}{c|}{41.76}                     & \multicolumn{1}{c|}{68.32}                     & \multicolumn{1}{c|}{3.98}                      & \multicolumn{1}{c|}{9.62}                      & 78.39                     & 31.69                      \\ \cline{3-20} 
                             &                            &                      & \Checkmark                  & \Checkmark                   &                       & \multicolumn{1}{c|}{8.49}                      & \multicolumn{1}{c|}{42.61}                     & \multicolumn{1}{c|}{48.82}                     & \multicolumn{1}{c|}{10.43}                     & \multicolumn{1}{c|}{38.29}                     & \multicolumn{1}{c|}{8.35}                      & \multicolumn{1}{c|}{16.73}                     & \multicolumn{1}{c|}{35.09}                     & \multicolumn{1}{c|}{35.83}                     & \multicolumn{1}{c|}{64.49}                     & \multicolumn{1}{c|}{8.44}                      & \multicolumn{1}{c|}{11.28}                     & 75.3                      & 31.08                      \\ \cline{3-20} 
\multirow{-12}{*}{ResNet-50} & \multirow{-6}{*}{MIMIC}    &                      & \Checkmark                  & \Checkmark                   & \Checkmark                   & \multicolumn{1}{c|}{9.13}                      & \multicolumn{1}{c|}{47.7}                      & \multicolumn{1}{c|}{54.91}                     & \multicolumn{1}{c|}{12.08}                     & \multicolumn{1}{c|}{43.83}                     & \multicolumn{1}{c|}{10.09}                     & \multicolumn{1}{c|}{16.42}                     & \multicolumn{1}{c|}{39.72}                     & \multicolumn{1}{c|}{43.18}                     & \multicolumn{1}{c|}{70.31}                     & \multicolumn{1}{c|}{10.33}                     & \multicolumn{1}{c|}{15.31}                     & 79.47                     & 34.81                      \\ 
\bottomrule
\end{tabular}
}
\end{table*}

\begin{table*}[]
\centering
\caption{Comparative experiments of using different datasets as augmented datasets. C and M denote the CheXpert and MIMIC datasets, respectively.}\label{table_mixda}
\scalebox{0.5}{
\begin{tabular}{c|c|c|c|ccccccccccccc|c}
\toprule
                            &                         &                       &                        & \multicolumn{13}{c|}{Disease Category}                                                                                                                                                                                                                                                                                                                                                                                                                                                    &                            \\ \cline{5-17}
\multirow{-2}{*}{Network}   & \multirow{-2}{*}{Train} & \multirow{-2}{*}{Aug} & \multirow{-2}{*}{Test} & \multicolumn{1}{c|}{{\color[HTML]{FE0000} EC}} & \multicolumn{1}{c|}{CA}    & \multicolumn{1}{c|}{LO}    & \multicolumn{1}{c|}{{\color[HTML]{FE0000} LL}} & \multicolumn{1}{c|}{ED}    & \multicolumn{1}{c|}{{\color[HTML]{FE0000} CO}} & \multicolumn{1}{c|}{{\color[HTML]{FE0000} PA}} & \multicolumn{1}{c|}{AT}    & \multicolumn{1}{c|}{PX}    & \multicolumn{1}{c|}{PE}    & \multicolumn{1}{c|}{{\color[HTML]{FE0000} PO}} & \multicolumn{1}{c|}{{\color[HTML]{FE0000} FE}} & SS    & \multirow{-2}{*}{F1 Score} \\ \hline
                            & C                       & -                     & C                      & \multicolumn{1}{c|}{7.22}                      & \multicolumn{1}{c|}{43.15} & \multicolumn{1}{c|}{69.73} & \multicolumn{1}{c|}{10.34}                     & \multicolumn{1}{c|}{45.8}  & \multicolumn{1}{c|}{4.2}                       & \multicolumn{1}{c|}{0.21}                      & \multicolumn{1}{c|}{19.72} & \multicolumn{1}{c|}{44.81} & \multicolumn{1}{c|}{73.31} & \multicolumn{1}{c|}{4.02}                      & \multicolumn{1}{c|}{11.9}                      & 78.06 & 31.72                      \\ \cline{2-18} 
                            & C                       & C                     & C                      & \multicolumn{1}{c|}{12.88}                     & \multicolumn{1}{c|}{38.7}  & \multicolumn{1}{c|}{71.32} & \multicolumn{1}{c|}{18.01}                     & \multicolumn{1}{c|}{54.14} & \multicolumn{1}{c|}{13.54}                     & \multicolumn{1}{c|}{9.31}                      & \multicolumn{1}{c|}{31.21} & \multicolumn{1}{c|}{39.7}  & \multicolumn{1}{c|}{71.68} & \multicolumn{1}{c|}{9.03}                      & \multicolumn{1}{c|}{18.62}                     & 74.83 & 35.61(+3.89)                      \\ \cline{2-18} 
                            & C                       & M                     & C                      & \multicolumn{1}{c|}{12.21}                     & \multicolumn{1}{c|}{36.39} & \multicolumn{1}{c|}{72.27} & \multicolumn{1}{c|}{16.85}                     & \multicolumn{1}{c|}{53.16} & \multicolumn{1}{c|}{12.59}                     & \multicolumn{1}{c|}{9.4}                       & \multicolumn{1}{c|}{31.48} & \multicolumn{1}{c|}{38.82} & \multicolumn{1}{c|}{72.31} & \multicolumn{1}{c|}{9.43}                      & \multicolumn{1}{c|}{17.49}                     & 72.43 & 34.98(+3.26)                      \\ \cline{2-18} 
                            & C                       & M+C                   & C                      & \multicolumn{1}{c|}{13.42}                     & \multicolumn{1}{c|}{39.4}  & \multicolumn{1}{c|}{72.55} & \multicolumn{1}{c|}{17.41}                     & \multicolumn{1}{c|}{56.4}  & \multicolumn{1}{c|}{14.77}                     & \multicolumn{1}{c|}{10.39}                     & \multicolumn{1}{c|}{32.76} & \multicolumn{1}{c|}{40.05} & \multicolumn{1}{c|}{73.12} & \multicolumn{1}{c|}{10.82}                     & \multicolumn{1}{c|}{17.6}                      & 75.72 & 36.49(+4.77)                      \\ \cline{2-18} 
                            & M                       & -                     & M                      & \multicolumn{1}{c|}{4.69}                      & \multicolumn{1}{c|}{47.61} & \multicolumn{1}{c|}{51.17} & \multicolumn{1}{c|}{9.28}                      & \multicolumn{1}{c|}{38.52} & \multicolumn{1}{c|}{1.41}                      & \multicolumn{1}{c|}{14.53}                     & \multicolumn{1}{c|}{38.89} & \multicolumn{1}{c|}{43.45} & \multicolumn{1}{c|}{63.41} & \multicolumn{1}{c|}{3.81}                      & \multicolumn{1}{c|}{12.03}                     & 75.57 & 31.11                      \\ \cline{2-18} 
                            & M                       & M                     & M                      & \multicolumn{1}{c|}{13.06}                     & \multicolumn{1}{c|}{47.69} & \multicolumn{1}{c|}{50.42} & \multicolumn{1}{c|}{14.77}                     & \multicolumn{1}{c|}{40.53} & \multicolumn{1}{c|}{12.48}                     & \multicolumn{1}{c|}{17.19}                     & \multicolumn{1}{c|}{39.41} & \multicolumn{1}{c|}{43.32} & \multicolumn{1}{c|}{64.27} & \multicolumn{1}{c|}{11.35}                     & \multicolumn{1}{c|}{14.16}                     & 72.24 & 33.91(+2.8)                      \\ \cline{2-18} 
                            & M                       & C                     & M                      & \multicolumn{1}{c|}{12.84}                     & \multicolumn{1}{c|}{47.31} & \multicolumn{1}{c|}{51.6}  & \multicolumn{1}{c|}{14.31}                     & \multicolumn{1}{c|}{41.28} & \multicolumn{1}{c|}{11.93}                     & \multicolumn{1}{c|}{16.79}                     & \multicolumn{1}{c|}{40.37} & \multicolumn{1}{c|}{42.14} & \multicolumn{1}{c|}{66.11} & \multicolumn{1}{c|}{10.24}                     & \multicolumn{1}{c|}{12.51}                     & 71.82 & 33.78(+2.67)                      \\ \cline{2-18} 
\multirow{-8}{*}{ResNet-50} & M                       & M+C                   & M                      & \multicolumn{1}{c|}{15.48}                     & \multicolumn{1}{c|}{48.12} & \multicolumn{1}{c|}{51.28} & \multicolumn{1}{c|}{15.03}                     & \multicolumn{1}{c|}{41.17} & \multicolumn{1}{c|}{13.25}                     & \multicolumn{1}{c|}{17.92}                     & \multicolumn{1}{c|}{40.15} & \multicolumn{1}{c|}{44.27} & \multicolumn{1}{c|}{64.79} & \multicolumn{1}{c|}{12.33}                     & \multicolumn{1}{c|}{14.8}                      & 72.63 & 34.71(+3.6)                       \\ \bottomrule
\end{tabular}
}
\end{table*}

\begin{figure}
\centering
\includegraphics[width=0.74\linewidth]{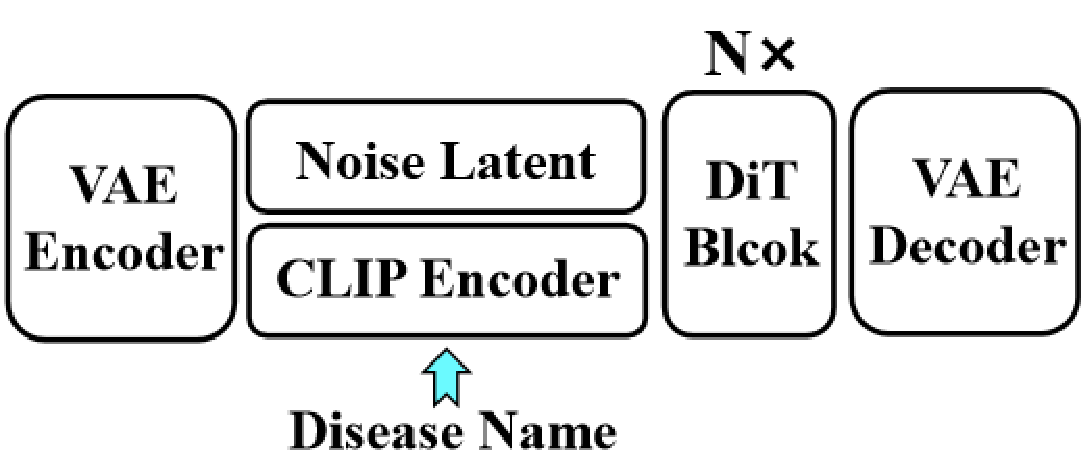}
\caption{The architecture of DiT variant.} \label{fig_variant}
\end{figure}

\noindent \textbf{Comparison to Caption-based Generation.} To further demonstrate the effectiveness of our proposed normal X-ray inpainting method, we designed a variant of DiT (see Figure~\ref{fig_variant}) by changing the original label input to the lesion category name. We used CLIP's text encoder~\cite{radford2021learning} to encode the lesion name as the network input and generated tail-class images by inputting the name of the tail-class lesion. The experimental results are shown in Table~\ref{table_caption}. As seen from the table, although the caption-based method also improves the performance of tail classes, its improvement is significantly lower than that of our inpainting-based method. Additionally, in terms of head-class lesions performance, the caption-based method does not outperform our method. We speculate that this result is due to the long-tail problem in the training dataset, which affects the training of DiT and, consequently, the quality of the generated data. The experimental results show that when the data used to train the diffusion model exhibits a long-tail distribution, the quality of the generated images decreases, thereby affecting the performance improvement of the network. In contrast, our proposed diffusion model, trained on normal X-ray data, demonstrates significant performance improvements due to the large volume of available training data.
\\

\noindent \textbf{Ablation Study.} In Table~\ref{table_loss_ablation}, we conducted ablation experiments to verify the effectiveness of the proposed progressive incremental learning strategy. 
Moreover, we introduced the focal loss~\cite{lin2017focal} to compare the performance improvement in tail classes. 
ResNet50 is taken as the verification network. From the table, we can see that although the classification accuracy achieved by the focal loss is higher than that of the cross-entropy loss for some head categories, it is lower than that of the cross-entropy loss in the tail categories. 
When PIL is not introduced into the network, the head classes will suffer from catastrophic forgetting due to the large amount of tail class data added to the training, which greatly reduces the performance of the head class. 
When PIL is added to the network training, the performance of the head classes is retained, and even improved for a few head classes. These results validated the effectiveness of our proposed PIL strategy.
\\

\noindent \textbf{Visualization of Inpainting Results.} In Figure~\ref{fig_inpating}, we visualize the inpainting images generated by the trained normal X-ray generator, where the guidance masks are extracted from the CAM maps. From the results generated in the last column, we can see that the edges of the generated area can be normally connected to the ribs of the original image, and the lung fibers in the original lesion area have also been restored to normal lung texture. These results demonstrate the powerful generation ability of the generator we trained.
\\

\begin{table}[]
\centering
\caption{Comparative experiments of different LLM models in LKM module.}\label{table_LLM}
\scalebox{0.39}{
\begin{tabular}{c|c|c|ccccccccccccc|c}
\toprule
                           &                       &                           & \multicolumn{13}{c|}{Disease Category}                                                                                                                                                                                                                                                                                                                                                                                                                                                    &                            \\ \cline{4-16}
\multirow{-2}{*}{Dataset}  & \multirow{-2}{*}{LLM} & \multirow{-2}{*}{Network} & \multicolumn{1}{c|}{{\color[HTML]{FE0000} EC}} & \multicolumn{1}{c|}{CA}    & \multicolumn{1}{c|}{LO}    & \multicolumn{1}{c|}{{\color[HTML]{FE0000} LL}} & \multicolumn{1}{c|}{ED}    & \multicolumn{1}{c|}{{\color[HTML]{FE0000} CO}} & \multicolumn{1}{c|}{{\color[HTML]{FE0000} PA}} & \multicolumn{1}{c|}{AT}    & \multicolumn{1}{c|}{PX}    & \multicolumn{1}{c|}{PE}    & \multicolumn{1}{c|}{{\color[HTML]{FE0000} PO}} & \multicolumn{1}{c|}{{\color[HTML]{FE0000} FE}} & SS    & \multirow{-2}{*}{F1 Score} \\ \hline
                           & -                     &                           & \multicolumn{1}{c|}{7.22}                      & \multicolumn{1}{c|}{43.15} & \multicolumn{1}{c|}{69.73} & \multicolumn{1}{c|}{10.34}                     & \multicolumn{1}{c|}{45.8}  & \multicolumn{1}{c|}{4.2}                       & \multicolumn{1}{c|}{0.21}                      & \multicolumn{1}{c|}{19.72} & \multicolumn{1}{c|}{44.81} & \multicolumn{1}{c|}{73.31} & \multicolumn{1}{c|}{4.02}                      & \multicolumn{1}{c|}{11.9}                      & 78.06 & 31.72                      \\ \cline{2-2} \cline{4-17} 
                           & GPT-4              &                           & \multicolumn{1}{c|}{12.88(+5.66)}              & \multicolumn{1}{c|}{38.7}  & \multicolumn{1}{c|}{71.32} & \multicolumn{1}{c|}{18.01(+7.67)}              & \multicolumn{1}{c|}{54.14} & \multicolumn{1}{c|}{13.54(+9.34)}              & \multicolumn{1}{c|}{9.31(+9.1)}                & \multicolumn{1}{c|}{31.21} & \multicolumn{1}{c|}{39.7}  & \multicolumn{1}{c|}{71.68} & \multicolumn{1}{c|}{9.03(+5.01)}               & \multicolumn{1}{c|}{18.62(+6.72)}              & 74.83 & 35.61               \\ \cline{2-2} \cline{4-17} 
                           & Grok                  &                           & \multicolumn{1}{c|}{11.52(+4.30)}              & \multicolumn{1}{c|}{40.25} & \multicolumn{1}{c|}{70.86} & \multicolumn{1}{c|}{16.83(+6.49)}              & \multicolumn{1}{c|}{52.37} & \multicolumn{1}{c|}{11.96(+7.76)}              & \multicolumn{1}{c|}{8.75(+8.54)}               & \multicolumn{1}{c|}{29.87} & \multicolumn{1}{c|}{41.35} & \multicolumn{1}{c|}{72.14} & \multicolumn{1}{c|}{8.27(+4.25)}               & \multicolumn{1}{c|}{17.95(+6.05)}              & 75.32 & 34.98               \\ \cline{2-2} \cline{4-17} 
                           & Doubao                &                           & \multicolumn{1}{c|}{13.25(+6.03)}              & \multicolumn{1}{c|}{37.92} & \multicolumn{1}{c|}{71.58} & \multicolumn{1}{c|}{17.46(+7.12)}              & \multicolumn{1}{c|}{53.89} & \multicolumn{1}{c|}{12.87(+8.67)}              & \multicolumn{1}{c|}{9.02(+8.81)}               & \multicolumn{1}{c|}{30.54} & \multicolumn{1}{c|}{39.16} & \multicolumn{1}{c|}{71.95} & \multicolumn{1}{c|}{8.76(+4.74)}               & \multicolumn{1}{c|}{18.23(+6.33)}              & 74.57 & 35.15              \\ \cline{2-2} \cline{4-17} 
                           & Gemini                &                           & \multicolumn{1}{c|}{12.13(+4.91)}              & \multicolumn{1}{c|}{39.58} & \multicolumn{1}{c|}{70.64} & \multicolumn{1}{c|}{17.12(+6.78)}              & \multicolumn{1}{c|}{52.96} & \multicolumn{1}{c|}{13.12(+8.92)}              & \multicolumn{1}{c|}{8.98(+8.77)}               & \multicolumn{1}{c|}{30.19} & \multicolumn{1}{c|}{40.82} & \multicolumn{1}{c|}{72.31} & \multicolumn{1}{c|}{8.54(+4.52)}               & \multicolumn{1}{c|}{17.68(+5.78)}              & 75.14 & 34.72               \\ \cline{2-2} \cline{4-17} 
                           & Llama                 &                           & \multicolumn{1}{c|}{11.87(+4.65)}              & \multicolumn{1}{c|}{38.36} & \multicolumn{1}{c|}{71.09} & \multicolumn{1}{c|}{16.59(+6.25)}              & \multicolumn{1}{c|}{51.74} & \multicolumn{1}{c|}{12.35(+8.15)}              & \multicolumn{1}{c|}{8.53(+8.32)}               & \multicolumn{1}{c|}{29.46} & \multicolumn{1}{c|}{41.68} & \multicolumn{1}{c|}{71.79} & \multicolumn{1}{c|}{7.98(+3.96)}               & \multicolumn{1}{c|}{17.32(+5.42)}              & 74.96 & 34.25               \\ \cline{2-2} \cline{4-17} 
                           & DeepSeek              &                           & \multicolumn{1}{c|}{12.56(+5.34)}              & \multicolumn{1}{c|}{39.12} & \multicolumn{1}{c|}{70.95} & \multicolumn{1}{c|}{17.78(+7.44)}              & \multicolumn{1}{c|}{53.28} & \multicolumn{1}{c|}{13.36(+9.16)}              & \multicolumn{1}{c|}{9.15(+8.94)}               & \multicolumn{1}{c|}{30.87} & \multicolumn{1}{c|}{39.94} & \multicolumn{1}{c|}{72.08} & \multicolumn{1}{c|}{8.89(+4.87)}               & \multicolumn{1}{c|}{18.45(+6.55)}              & 75.03 & 35.38               \\ \cline{2-2} \cline{4-17} 
                           & Qwen                  &                           & \multicolumn{1}{c|}{11.98(+4.76)}              & \multicolumn{1}{c|}{38.85} & \multicolumn{1}{c|}{71.23} & \multicolumn{1}{c|}{16.97(+6.63)}              & \multicolumn{1}{c|}{52.15} & \multicolumn{1}{c|}{12.68(+8.48)}              & \multicolumn{1}{c|}{8.86(+8.65)}               & \multicolumn{1}{c|}{29.73} & \multicolumn{1}{c|}{41.19} & \multicolumn{1}{c|}{71.86} & \multicolumn{1}{c|}{8.15(+4.13)}               & \multicolumn{1}{c|}{17.82(+5.92)}              & 74.79 & 34.51               \\ \cline{2-2} \cline{4-17} 
\multirow{-9}{*}{CheXpert} & ChatGLM               &                           & \multicolumn{1}{c|}{12.34(+5.12)}              & \multicolumn{1}{c|}{39.76} & \multicolumn{1}{c|}{70.78} & \multicolumn{1}{c|}{17.25(+6.91)}              & \multicolumn{1}{c|}{53.02} & \multicolumn{1}{c|}{12.99(+8.79)}              & \multicolumn{1}{c|}{8.67(+8.46)}               & \multicolumn{1}{c|}{30.35} & \multicolumn{1}{c|}{40.57} & \multicolumn{1}{c|}{72.24} & \multicolumn{1}{c|}{8.42(+4.40)}               & \multicolumn{1}{c|}{18.11(+6.21)}              & 75.26 & 34.89               \\ \cline{1-2} \cline{4-17} 
                           & -                     &                           & \multicolumn{1}{c|}{4.69}                      & \multicolumn{1}{c|}{47.61} & \multicolumn{1}{c|}{51.17} & \multicolumn{1}{c|}{9.28}                      & \multicolumn{1}{c|}{38.52} & \multicolumn{1}{c|}{1.41}                      & \multicolumn{1}{c|}{14.53}                     & \multicolumn{1}{c|}{38.89} & \multicolumn{1}{c|}{43.35} & \multicolumn{1}{c|}{63.41} & \multicolumn{1}{c|}{3.81}                      & \multicolumn{1}{c|}{12.03}                     & 75.57 & 31.11                      \\ \cline{2-2} \cline{4-17} 
                           & GPT-4              &                           & \multicolumn{1}{c|}{13.06(+8.37)}              & \multicolumn{1}{c|}{47.69} & \multicolumn{1}{c|}{50.42} & \multicolumn{1}{c|}{14.77(+5.49)}              & \multicolumn{1}{c|}{40.53} & \multicolumn{1}{c|}{12.48(+11.07)}             & \multicolumn{1}{c|}{17.19(+2.66)}              & \multicolumn{1}{c|}{39.41} & \multicolumn{1}{c|}{43.32} & \multicolumn{1}{c|}{64.27} & \multicolumn{1}{c|}{11.35(+7.54)}              & \multicolumn{1}{c|}{14.16(+2.13)}              & 72.24 & 33.91                \\ \cline{2-2} \cline{4-17} 
                           & Grok                  &                           & \multicolumn{1}{c|}{12.35(+7.66)}              & \multicolumn{1}{c|}{46.98} & \multicolumn{1}{c|}{50.76} & \multicolumn{1}{c|}{13.92(+4.64)}              & \multicolumn{1}{c|}{39.87} & \multicolumn{1}{c|}{11.75(+10.34)}             & \multicolumn{1}{c|}{16.58(+2.05)}              & \multicolumn{1}{c|}{38.96} & \multicolumn{1}{c|}{42.87} & \multicolumn{1}{c|}{63.95} & \multicolumn{1}{c|}{10.72(+6.91)}              & \multicolumn{1}{c|}{13.78(+1.75)}              & 72.89 & 33.15               \\ \cline{2-2} \cline{4-17} 
                           & Doubao                &                           & \multicolumn{1}{c|}{12.78(+8.09)}              & \multicolumn{1}{c|}{47.32} & \multicolumn{1}{c|}{50.59} & \multicolumn{1}{c|}{14.35(+5.07)}              & \multicolumn{1}{c|}{40.21} & \multicolumn{1}{c|}{12.16(+10.75)}             & \multicolumn{1}{c|}{16.93(+2.40)}              & \multicolumn{1}{c|}{39.18} & \multicolumn{1}{c|}{43.05} & \multicolumn{1}{c|}{64.12} & \multicolumn{1}{c|}{11.08(+7.27)}              & \multicolumn{1}{c|}{14.03(+1.99)}              & 72.56 & 33.42               \\ \cline{2-2} \cline{4-17} 
                           & Gemini                &                           & \multicolumn{1}{c|}{12.19(+7.50)}              & \multicolumn{1}{c|}{47.15} & \multicolumn{1}{c|}{50.68} & \multicolumn{1}{c|}{13.67(+4.39)}              & \multicolumn{1}{c|}{39.65} & \multicolumn{1}{c|}{11.92(+10.51)}             & \multicolumn{1}{c|}{16.34(+1.81)}              & \multicolumn{1}{c|}{38.83} & \multicolumn{1}{c|}{42.96} & \multicolumn{1}{c|}{63.87} & \multicolumn{1}{c|}{10.54(+6.73)}              & \multicolumn{1}{c|}{13.56(+1.53)}              & 73.01 & 32.98               \\ \cline{2-2} \cline{4-17} 
                           & Llama                 &                           & \multicolumn{1}{c|}{11.87(+7.18)}              & \multicolumn{1}{c|}{46.72} & \multicolumn{1}{c|}{50.89} & \multicolumn{1}{c|}{13.42(+4.14)}              & \multicolumn{1}{c|}{39.34} & \multicolumn{1}{c|}{11.58(+10.17)}             & \multicolumn{1}{c|}{16.12(+1.59)}              & \multicolumn{1}{c|}{38.67} & \multicolumn{1}{c|}{43.18} & \multicolumn{1}{c|}{63.79} & \multicolumn{1}{c|}{10.36(+6.55)}              & \multicolumn{1}{c|}{13.39(+1.36)}              & 73.25 & 32.65               \\ \cline{2-2} \cline{4-17} 
                           & DeepSeek              &                           & \multicolumn{1}{c|}{12.89(+8.20)}              & \multicolumn{1}{c|}{47.46} & \multicolumn{1}{c|}{50.53} & \multicolumn{1}{c|}{14.52(+5.24)}              & \multicolumn{1}{c|}{40.38} & \multicolumn{1}{c|}{12.34(+10.93)}             & \multicolumn{1}{c|}{17.05(+2.52)}              & \multicolumn{1}{c|}{39.29} & \multicolumn{1}{c|}{43.21} & \multicolumn{1}{c|}{64.19} & \multicolumn{1}{c|}{11.23(+7.42)}              & \multicolumn{1}{c|}{14.25(+2.22)}              & 72.43 & 33.68               \\ \cline{2-2} \cline{4-17} 
                           & Qwen                  &                           & \multicolumn{1}{c|}{12.03(+7.34)}              & \multicolumn{1}{c|}{46.89} & \multicolumn{1}{c|}{50.72} & \multicolumn{1}{c|}{13.75(+4.47)}              & \multicolumn{1}{c|}{39.51} & \multicolumn{1}{c|}{11.84(+10.43)}             & \multicolumn{1}{c|}{16.47(+1.94)}              & \multicolumn{1}{c|}{38.75} & \multicolumn{1}{c|}{43.09} & \multicolumn{1}{c|}{63.92} & \multicolumn{1}{c|}{10.68(+6.87)}              & \multicolumn{1}{c|}{13.69(+1.66)}              & 72.95 & 32.87               \\ \cline{2-2} \cline{4-17} 
\multirow{-9}{*}{MIMIC}    & ChatGLM               & \multirow{-18}{*}{R50}    & \multicolumn{1}{c|}{12.56(+7.87)}              & \multicolumn{1}{c|}{47.21} & \multicolumn{1}{c|}{50.61} & \multicolumn{1}{c|}{14.18(+4.90)}              & \multicolumn{1}{c|}{40.07} & \multicolumn{1}{c|}{12.05(+10.64)}             & \multicolumn{1}{c|}{16.79(+2.26)}              & \multicolumn{1}{c|}{39.04} & \multicolumn{1}{c|}{43.15} & \multicolumn{1}{c|}{64.05} & \multicolumn{1}{c|}{10.95(+7.14)}              & \multicolumn{1}{c|}{13.92(+1.89)}              & 72.78 & 33.29              \\ \bottomrule
\end{tabular}
}
\end{table}

\noindent \textbf{Comparison of Different LLM Models in LKG.} 
In Table~\ref{table_LLM}, we conducted a comparative analysis of different Large Language Models (LLMs) integrated into the LKG module. 
We use the ResNet-50 as the evaluation network.
The experimental results demonstrate that incorporating lung-related pathology knowledge from any of these LLMs improves the module's performance. 
Although GPT-4 achieves the highest overall F1 scores (35.61\% on CheXpert and 33.91\% on MIMIC-CXR), other LLMs exhibit comparable performance and even surpass GPT-4 in certain classes.\\

\noindent \textbf{Mixed Dataset Training.} To further demonstrate that the normal X-ray generator we trained has strong generation capabilities, we conducted mixed inpainting data training on MIMIC-CXR and CheXpert datasets. The experimental results are shown in Table~\ref{table_mixda}. From the table, we can see that when the augmented dataset uses inpainting data other than the current dataset as augmented data for training, the network performance can still be improved, by 3.26\% and 2.67\% on CheXpert and MIMIC, respectively. When the augmented dataset includes both datasets, the network performance can be further improved, by 4.77\% and 3.6\%, respectively. 
These experimental results show that the generator we trained is compatible with different data sources and generates better tail class data to improve the performance of the network.

\section{Conclusion} 
In this paper, we proposed a novel data synthesis pipeline to augment tail-class lesions with commonly sufficient normal X-rays. 
By inpainting the head-class lesions area in the X-ray into normal lung texture and retaining the tail-class lesions area, new tailed data is constructed to improve the performance of the tail classes. 
A large language model knowledge guidance (LKG) module and a progressive incremental learning (PIL) strategy were proposed to stabilize the inpaitining fine-tuning. 
Extensive experiments on public lung datasets MIMIC and CheXpert demonstrated that the proposed method can achieve state-of-the-art performance. 




\bibliographystyle{elsarticle-num.bst}
\bibliography{ref}

\end{document}